\pdfoutput=1 
\documentclass[10pt,twocolumn,letterpaper]{article}
\usepackage{cvpr}
\usepackage{times}
\usepackage{epsfig}
\usepackage{amsmath}
\usepackage{amssymb}
\usepackage{graphicx}
\usepackage{subcaption}
\usepackage{multirow}

\newcommand{\comment}[1]{}
\newcommand{\dataset}{Agriculture-Vision}

\usepackage[pagebackref=true,breaklinks=true,letterpaper=true,colorlinks,bookmarks=false]{hyperref}

\cvprfinalcopy 

\def\cvprPaperID{0346} 

\ifcvprfinal\fi
\begin{document}

\title{\dataset{}: A Large Aerial Image Database for \\ Agricultural Pattern Analysis
}

\author{Mang Tik Chiu$^{1*}$, Xingqian Xu$^{1}$\thanks{~indicates joint first author. For more information on our database and other related efforts in Agriculture-Vision, please visit our CVPR 2020 workshop and challenge website {\color{blue} https://www.agriculture-vision.com}.}~~, Yunchao Wei$^{1}$, Zilong Huang$^1$, 
\\Alexander Schwing$^1$, Robert Brunner$^{1}$, Hrant Khachatrian$^2$, Hovnatan Karapetyan$^2$, \\Ivan Dozier$^2$, Greg Rose$^2$, David Wilson$^2$, Adrian Tudor$^2$, Naira Hovakimyan$^{2,1}$, \\Thomas S. Huang$^1$, Honghui Shi$^{3,1}$
\\
\\
$^1$UIUC, $^2$Intelinair, $^3$University of Oregon}

\maketitle

\begin{abstract}
    The success of deep learning in visual recognition tasks has driven advancements in multiple fields of research. Particularly, increasing attention has been drawn towards its application in agriculture. Nevertheless, while visual pattern recognition on farmlands carries enormous economic values, little progress has been made to merge computer vision and crop sciences due to the lack of suitable agricultural image datasets. Meanwhile, problems in agriculture also pose new challenges in computer vision. For example, semantic segmentation of aerial farmland images requires inference over extremely large-size images with extreme annotation sparsity. These challenges are not present in most of the common object datasets, and we show that they are more challenging than many other aerial image datasets. To encourage research in computer vision for agriculture, we present \textbf{\dataset{}}: a large-scale aerial farmland image dataset for semantic segmentation of agricultural patterns.
We collected $94,986$ high-quality aerial images from $3,432$ farmlands across the US, where each image consists of RGB and Near-infrared (NIR) channels with resolution as high as 10 cm per pixel. We annotate nine types of field anomaly patterns that are most important to farmers. As a pilot study of aerial agricultural semantic segmentation, we perform comprehensive experiments using popular semantic segmentation models; we also propose an effective model designed for aerial agricultural pattern recognition. Our experiments demonstrate several challenges \dataset{} poses to both the computer vision and agriculture communities. Future versions of this dataset will include even more aerial images, anomaly patterns and image channels.

\vspace{-5mm}
\end{abstract}

\section{Introduction}
Since the introduction of ImageNet~\cite{deng2009imagenet}, a large-scale image classification dataset, research in computer vision and pattern recognition using deep neural nets has seen unprecedented development~\cite{krizhevsky2012imagenet, he2016deep, szegedy2015going, simonyan2014very, huang2019convolutional}. Deep neural networks based algorithms have proven to be effective across multiple domains such as medicine and astronomy~\cite{larson2017performance,aniyan2017classifying,yu2020foal}, across multiple datasets~\cite{guo2019spottune,wang2020differential,fu2019self}, across different computer vision tasks~\cite{yu2017computed,jiao2019geometry, yu2019novel, cheng2018revisiting2, cheng2019bottom, cheng2018decoupled, shi2018geometry, qian2019weakly,yu2020foal} and across different numerical precision and hardware architectures~\cite{yu2019anyprecision, zhang2019skynet}. 
However, progress of visual pattern recognition in agriculture, one of the fundamental aspects of the human race, has been relatively slow~\cite{kamilaris2018deep}. This is partially due to the lack of relevant datasets that encourage the study of agricultural imagery and visual patterns, which poses many distinctive characteristics.

A major direction of visual recognition in agriculture is aerial image semantic segmentation. Solving this problem is important because it has tremendous economic potential. Specifically, efficient algorithms for detecting field conditions enable timely actions to prevent major losses or to increase potential yield throughout the growing season. However, this is much more challenging compared to typical semantic segmentation tasks on other aerial image datasets. For example, to segment weed patterns in aerial farmland images, the algorithm must be able to identify sparse weed clusters of vastly different shapes and coverages. In addition, some of these aerial images have sizes exceeding $20000\times30000$ pixels, these images pose a huge problem for end-to-end segmentation in terms of computation power and memory consumption. Agricultural data are also inherently multi-modal, where information such as field temperature and near-infrared signal are essential for determining field conditions. These properties deviate from those of conventional semantic segmentation tasks, thus reducing their applicability to this area of research.

\begin{table*}[bt!]
\centering
\def\arraystretch{1.1}   
\resizebox{0.98\textwidth}{!}{%
    \begin{tabular}{r|c|c|c|c|c|c|c|c}
    \hline
    \multicolumn{1}{c|}{Dataset} & \# Images & \# Classes & \# Labels & Tasks & \begin{tabular}[x]{@{}c@{}}Image Size\\(pixels)\end{tabular}  & \# Pixels & Channels & \begin{tabular}[x]{@{}c@{}}Resolution\\(GSD)\end{tabular} \\ \hline \hline
    
    \multicolumn{1}{l|}{\textit{Aerial images}} &&&&&&& \\
    
    Inria Aerial Image \cite{maggiori2017dataset} & 180 & 2 & 180 & seg. & $5000\times5000$ & 4.5B & RGB & 30 cm/px\\ 
    
    DOTA \cite{xia2018dota} & 2,806 & 14 & 188,282 & det. & $\leq4000\times4000$ & 44.9B & RGB & various \\ 
    
    iSAID \cite{waqas2019isaid} & 2,806 & 15 & 655,451 & seg. & $\leq4000\times4000$ & 44.9B & RGB & various \\ 
    
    AID \cite{xia2017aid} & 10,000 & 30 & 10,000 & cls.  & $600\times600$ & 3.6B & RGB & 50-800 cm/px \\ 
    
    DeepGlobe Building \cite{demir2018deepglobe} & 24,586 & 2 & 302,701 & det. / seg. & $650\times650$ & 10.4B & 9 bands & 31-124 cm/px \\
   
    EuroSAT \cite{helber2017eurosat} & 27,000 & 10 & 27,000 & cls.  & $256\times256$ & 1.77B & 13 Bands & 30 cm/px \\
    
    SAT-4 \cite{basu2015deepsat} & 500,000 & 4 & 500,000 & cls.  & $28\times28$ & 0.39B & RGB, NIR & 600 cm/px\\ 
    
    SAT-6 \cite{basu2015deepsat} & 405,000 & 6 & 405,000 & cls.  & $28\times28$ & 0.32B & RGB, NIR & 600 cm/px \\ \hline
    
    \multicolumn{1}{l|}{\textit{Agricultural images}} &&&&&&& \\
    
    Crop/Weed discrimination \cite{haug2014crop} & 60 & 2 & 494 & seg. & $1296\times966$ & 0.08B & RGB & N/A \\
    
    Sensefly Crop Field \cite{sensefly} & 5,260 & N/A & N/A & N/A & N/A & N/A & NRG, Red edge & 12.13 cm/px \\ 
    
    DeepWeeds \cite{olsen2019deepweeds} & 17,509 & 1$^\dagger$ & 17,509 & cls. & $1920\times1200$ & 40.3B & RGB & N/A \\ 
    
    \dataset{} (ours) & \textbf{94,986} & \textbf{9} & \textbf{169,086} & seg. & $512\times512$ & 22.6B & \textbf{RGB, NIR} & \textbf{10/15/20 cm/px}\\ 
    \hline
    \end{tabular}%
  }
\vspace{-2mm}
\begin{flushleft}
\scriptsize 
  $\dagger$ DeepWeeds has only weed annotations at image-level, but there are 8 sub-categories of weeds.
\end{flushleft}
\vspace{-3mm}
  \caption{This table shows the statistics from other datasets. All datasets are compared on number of images, categories, annotations, image size, pixel numbers and color channels. If it is an aerial image dataset, we also provide the ground sample resolution (GSD). ``cls.'', ``det.'' and ``seg.'' stand for classification, detection and segmentation respectively.}
  \label{table:related_dataset}
\vspace{-5mm}
\end{table*}

To encourage research on this challenging task, we present \dataset{}, a large-scale and high-quality dataset of aerial farmland images for advancing studies of agricultural semantic segmentation. We collected images throughout the growing seasons at numerous farming locations in the US, where several important field patterns were annotated by agronomy experts.

\dataset{} differs significantly from other image datasets in the following aspects: (1) unprecedented aerial image resolutions up to 10 cm per pixel (cm/px); (2) multiple aligned image channels beyond RGB; (3) challenging annotations of multiple agricultural anomaly patterns; (4) precise annotations from professional agronomists with a strict quality assurance process; and (5) large size and shape variations of annotations. These features make \dataset{} a unique image dataset that poses new challenges for semantic segmentation in aerial agricultural images.

Our main contributions are summarized as follows:
\begin{itemize}
    \vspace{-2mm}
    \item We introduce a large-scale and high quality aerial agricultural image database for advancing research in agricultural pattern analysis and semantic segmentation.
    \vspace{-2mm}
    \item We perform a pilot study with extensive experiments on the proposed database and provide a baseline for semantic segmentation using deep learning approaches to encourage further research.
    \vspace{-2mm}
\end{itemize}

\section{Related Work}
\label{sec:related_work}
Most segmentation datasets primarily focus on common objects or street views. For example, Pascal VOC~\cite{everingham2010pascal}, MS-COCO~\cite{lin2014microsoft} and ADE20K~\cite{zhou2017scene} segmentation datasets respectively consist of 20, 91 and 150 daily object categories such as airplane, person, computer, etc. The Cityscapes dataset~\cite{cordts2016cityscapes}, where dense annotations of street scenes are available, opened up research directions in street-view scene parsing and encouraged more research efforts in this area.

Aerial image visual recognition has also gained increasing attention. Unlike daily scenes, aerial images are often significantly larger in image sizes. For example, the DOTA dataset~\cite{xia2018dota} contains images with sizes up to $4000\times4000$ pixels, which are significantly larger than those in common object datasets at around $500\times500$ pixels. Yet, aerial images are often of much lower resolutions. Precisely, the CVPR DeepGlobe2018 Building Extraction Challenge~\cite{demir2018deepglobe} uses aerial images at a resolution of 31 cm/px or lower. As a result, finer object details such as shape and texture are lost and have to be omitted in later studies.

Table~\ref{table:related_dataset} summarizes the statistics of the most related datasets, including those of aerial images and agricultural images. As can be seen from the table, there has been an apparent lack of large-scale aerial agricultural image databases, which, in some sense, hinders agricultural visual recognition research from rapid growth as evidenced for common images~\cite{mishkin2017systematic}.

Meanwhile, many agricultural studies have proposed solutions to extract meaningful information through images. These papers cover numerous subtopics, such as spectral analysis on land and crops~\cite{zhao2019use, lebourgeois2008can, hunt2010acquisition, kelcey2012sensor}, aerial device photogrammetry~\cite{haala2011performance, laliberte2011image}, color indices and low-level image feature analysis~\cite{unnikrishnan2019deep, ramanath2019ndvi, gitelson2002novel, woebbecke1995color, el2000weed}, as well as integrated image processing systems~\cite{laliberte2011image, lamm2002precision}. One popular approach in analyzing agricultural images is to use geo-color indices such as the Normalized-Difference-Vegetation-Index (NDVI) and Excess-Green-Index (ExG). These indices have high correlation with land information such as water~\cite{zarco2013pri} and plantations~\cite{marchant2001evaluation}. Besides, recent papers in computer vision have been eminently motivated by deep convolutional neural networks (DCNN)~\cite{krizhevsky2012imagenet}. DCNN is also in the spotlight in agricultural vision problems such as land cover classification~\cite{lu2017cultivated} and weed detection~\cite{milioto2017real}. In a similar work~\cite{lu2017cultivated}, Lu et. al. collected aerial images using an EOS 5D camera at 650m and 500m above ground in Penzhou and Guanghan County, Sichuan, China. They labeled cultivated land vs. background using a three-layer CNN model. In another recent work~\cite{rebetez2016augmenting}, Rebetez et. al. utilized an experimental farmland dataset conducted by the Swiss Confederation's Agroscope research center and proposed a DCNN-HistNN hybrid model to categorize plant species on a pixel-level. Nevertheless, since their datasets are limited in scale and their research models are outdated, both works fail to fuse state-of-the-art deep learning approaches in agricultural applications in the long run.

\section{The \dataset{} Dataset}
\dataset{} aims to be a publicly available large-scale aerial agricultural image dataset that is high-resolution, multi-band, and with multiple types of patterns annotated by agronomy experts. In its current stage, we have captured 3,432 farmland images with nine types of annotations: double plant, drydown, endrow, nutrient deficiency, planter skip, storm damage, water, waterway and weed cluster. All of these patterns have substantial impacts on field conditions and the final yield. These farmland images were captured between 2017 and 2019 across multiple growing seasons in numerous farming locations in the US. The proposed \dataset{} dataset contains 94,986 images sampled from these farmlands. In this section, we describe the details on how we construct the \dataset{} dataset, including image acquisition, preprocessing, pattern annotation, and finally image sample generation.

\subsection{Field Image Acquisition}

Farmland images in the \dataset{} dataset were captured by specialized mounted cameras on aerial vehicles flown over numerous fields in the US, which primarily consist of corn and soybean fields around Illinois and Iowa. 
All images in the current version of \dataset{} were collected from the growing seasons between 2017 and 2019. Each field image contains four color channels: Near-infrared (NIR), Red, Green and Blue. 

\begin{table}[h!]
\centering
\resizebox{0.99\linewidth}{!}{%
  \begin{tabular}{c|c|c|c|c}
    \hline
    Year & Channel & Resolution & Description & Camera \\ \hline \hline
    2017 & N, R, G, B & 15cm/px & Narrow band & $2\times$Canon SLR \\ \hline
    \multirow{2}{*}{2018} & N, R, G & 10cm/px & Narrow band & $2\times$Nikon D850 \\
    & B & 20cm/px & Wide band & $1\times$Nikon D800E \\ 
    \hline
    2019 & N, R, G, B & 10cm/px & Narrow band & WAMS \\
    \hline
  \end{tabular}%
  }
  \newline
  \caption{Camera settings for capturing the 4-channel field images: Near-infrared (N), Red (R), Green (G) and Blue (B). The Blue channel images captured in 2018 are scaled up to align with the NRG images.}
  \label{table:camera_spec}
  \vspace{-5mm}
\end{table}

The camera settings for capturing farmland images are shown in Table~\ref{table:camera_spec}.
Farmland images in 2017 were taken with two aligned Canon SLR cameras, where one captures RGB images and the other captures only the NIR channel. For farmland images in 2018, the NIR, Red and Green (NRG) channels were taken using two Nikon D850 cameras to enable 10 cm/px resolution. Custom filters were used to capture near-infrared instead of the blue channel. Meanwhile, the separate Blue channel images were captured using one Nikon D800E at 20 cm/px resolution, which were then scaled up to align with the corresponding NRG images. Farmland images in 2019 were captured using a proprietary Wide Area Multi-Spectral System (WAMS) commonly used for remote sensing. The WAMS captures all four channels simultaneously at 10 cm/px resolution. Note that compared to other aerial image datasets in Table~\ref{table:related_dataset}, our dataset contains images in resolutions higher than all others. 

\begin{figure*}[ht!]
    \centering
    \includegraphics[width=\linewidth]{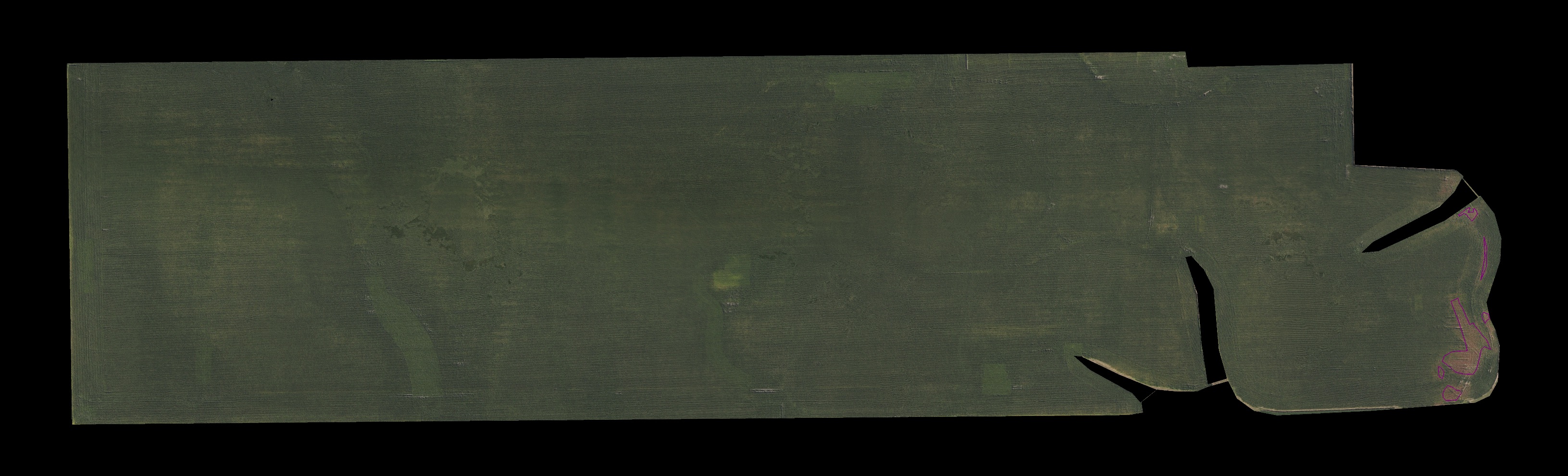}
    \vspace{-0.3cm}
    \caption{Visualization of an aerial farmland image before sub-sampling. This image (including invalid areas, shown in black) has a size of $10875\times3303$ pxiels and only contains drydown annotations at the rightmost region. Due to the large image size and sparse annotation, training semantic segmentation models on entire images is impractical and inefficient.}
    \label{fig:field_vis}
    \vspace{-3mm}
\end{figure*}

\subsection{Farmland image preprocessing}

Farmland images captured in 2017 were already stored in regular pixel values between 0 and 255, while those captured in 2018 and 2019 were initially stored in camera raw pixel format. Following the conventional method for normalizing agricultural images, for each of the four channels in one field image, we first compute the $5^{th}$ and $95^{th}$ percentile pixel values, then clip all pixel values in the image by a lower bound and an upper bound:
\begin{equation}
\label{eq:pix_low_up}
\begin{aligned}
    V_{lower} &= max(0, p_5 - 0.4 \times (p_{95} - p_5)) \\
    V_{upper} &= min(255, p_{95} + 0.4 \times (p_{95} - p_5))
\end{aligned}
\end{equation}
where $V_{lower}$, $V_{upper}$ stand for lower and upper bound of pixel values respectively, $p_5$ and $p_{95}$ stand for the $5^{th}$ and $95^{th}$ percentile respectively.

Note that farmland images may contain invalid areas, which were initially marked with a special pixel value. Therefore, we exclude these invalid areas when computing pixel percentiles for images in 2018 and 2019.

To intuitively visualize each field image and prepare for later experiments, we separate the four channels into a regular RGB image and an additional single-channel NIR image, and store them as two JPG images.

\subsection{Annotations}

All annotations in \dataset{} were labeled by five annotators trained by expert agronomists through a commercial software. Annotated patterns were then reviewed by the agronomists, where unsatisfactory annotations were improved. The software provides visualizations of several image channels and vegetation indices, including RGB, NIR and NDVI, where NDVI can be derived from the Red and NIR channel by:

\begin{equation}
\begin{aligned}
    NDVI &= \frac{NIR-RED}{NIR+RED}
\end{aligned}
\label{eq:ndvi_calc}
\end{equation}


\subsection{Image sample generation}

Unprocessed farmland images have extremely large image sizes. 
For instance, Figure~\ref{fig:field_vis} shows one field image with a size of $10875\times3303$ pixels. In fact, the largest field image we collected is $33571\times24351$ pixels in size. This poses significant challenges to deep network training in terms of computation time and memory consumption. In addition, Figure~\ref{fig:field_vis} also shows the sparsity of some annotations. This means training a segmentation model on the entire image for these patterns would be very inefficient, and would very possibly yield suboptimal results.

On the other hand, unlike common objects, visual appearances of anomaly patterns in aerial farmland images are preserved under image sub-sampling methods such as flipping and cropping. This is because these patterns represent \textit{regions} of the anomalies instead of individual objects. As a result, we can sample image patches from these large farmland images by cropping around annotated regions in the image. This simultaneously improves data efficiency, since the proportion of annotated pixels is increased.

Motivated by the above reasons, we construct the \dataset{} dataset by cropping annotations with a window size of $512\times512$ pixels. For field patterns smaller than the window size, we simply crop the region centered at the annotation. For field patterns larger than the window size, we employ a non-overlapping sliding window technique to cover the entirety of the annotation. Note that we discard images covered by more than 90\% of annotations, such that all images retain sufficient context information.

\begin{figure}[h!]
\centering
\includegraphics[width=0.65\linewidth]{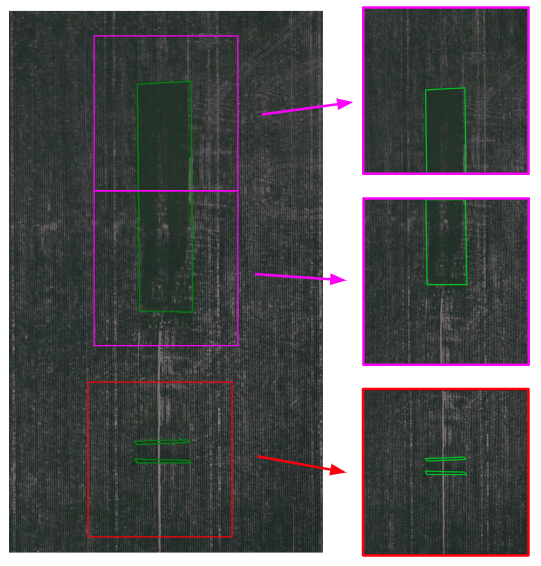}
\vspace{-2mm}
\caption{This figure illustrates our field image patch generation method for AgriVision. For annotations smaller than $512\times512$ pixels, we crop the image by a single window around the annotation center (shown in red). For larger annotations, we use multiple non-overlapping windows to cover the entire annotation (shown in purple). Note that the bottom two polygons are enclosed by just one window.}
\label{fig:image_sample_generation}
\vspace{-5mm}
\end{figure}

In many cases, multiple small annotations are located near each other. Generating one image patch for every annotation would lead to severe re-sampling of those field regions, which causes biases in the dataset. To alleviate the issue, if two image patches have an Intersection-over-Union of over 30\%, we discard the one with fewer pixels annotated as field patterns. When cropping large annotations using a sliding window, we also discard any image patches with only background pixels. A visualization of our sample generation method is illustrated in Figure~\ref{fig:image_sample_generation}, and some images in the final \dataset{} dataset are shown in Figure~\ref{fig:crop_vis}.

\vspace{-2mm}
\newcommand{\img}[1]{
    \includegraphics[width=\linewidth]{IMAGES/crop_vis/#1_rgb.jpg}\\
    \vspace{0.1cm}
    \includegraphics[width=\linewidth]{IMAGES/crop_vis/#1_nrg.jpg}
}

\newcommand{\imgbox}[2]{
    \begin{subfigure}{#1\linewidth}
    \centering
    #2
    \end{subfigure}
}

\begin{figure*}[ht!]
\centering
\vspace{0.1cm}
    \imgbox{0.32}{
        \imgbox{0.48}{\img{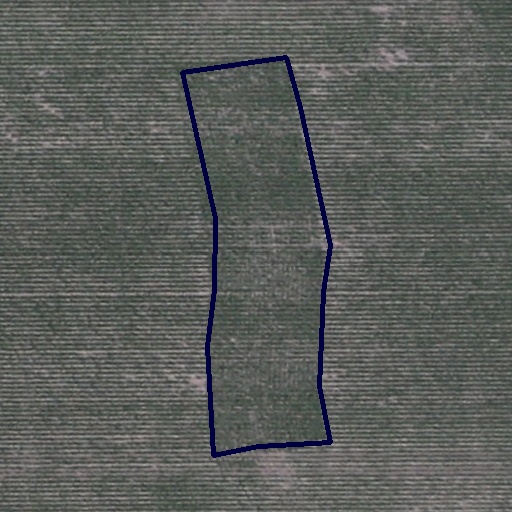}}%
        \imgbox{0.48}{\img{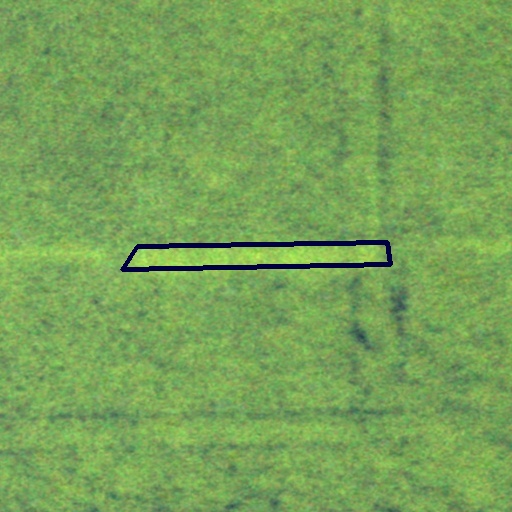}}
        \vspace{-0.5cm}
        \caption{Double plant}
        \vspace{1cm}
    }%
    \imgbox{0.32}{
        \imgbox{0.48}{\img{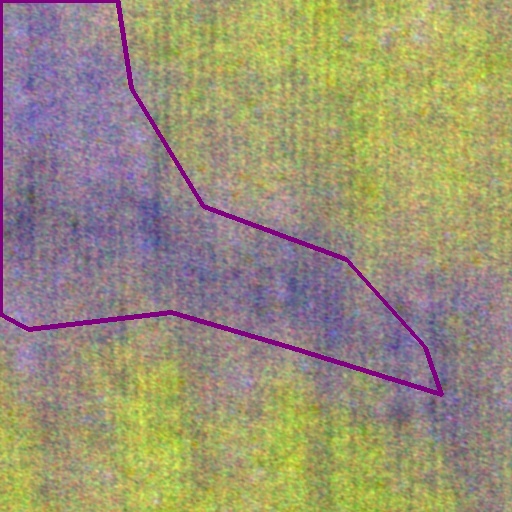}}%
        \imgbox{0.48}{\img{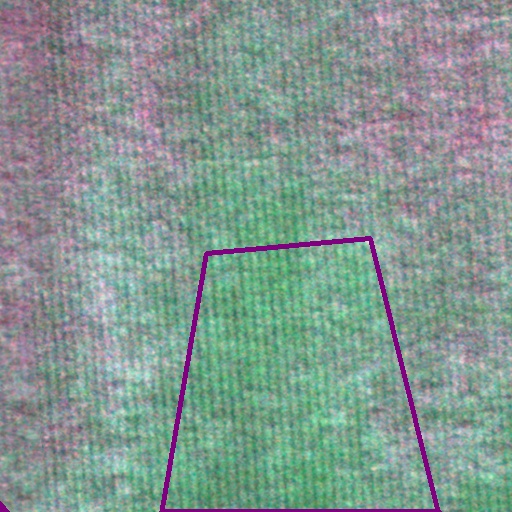}}
        \vspace{-0.5cm}
        \caption{Drydown}
        \vspace{1cm}
    }%
    \imgbox{0.32}{
        \imgbox{0.48}{\img{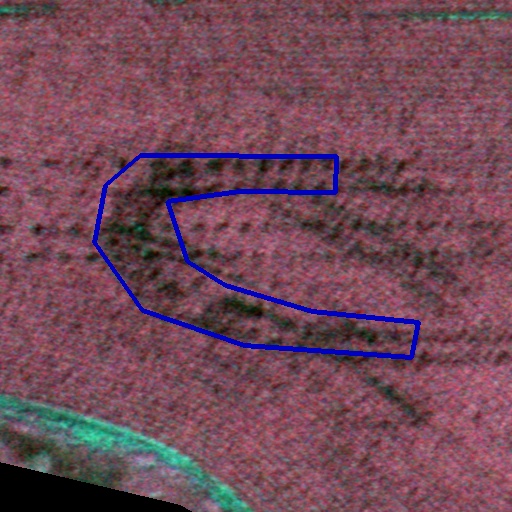}}%
        \imgbox{0.48}{\img{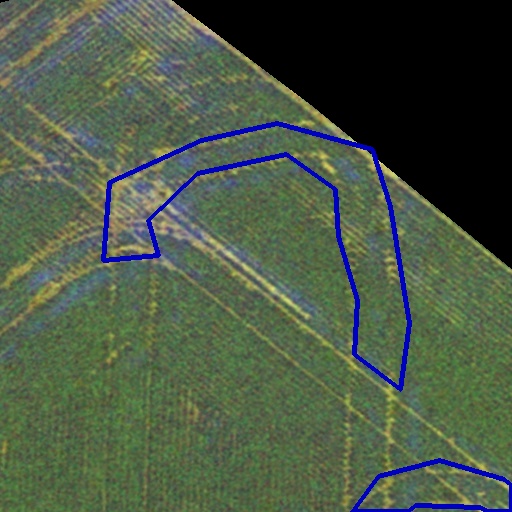}}
        \vspace{-0.5cm}
        \caption{Endrow}
        \vspace{1cm}
    }
    \imgbox{0.32}{
        \imgbox{0.48}{\img{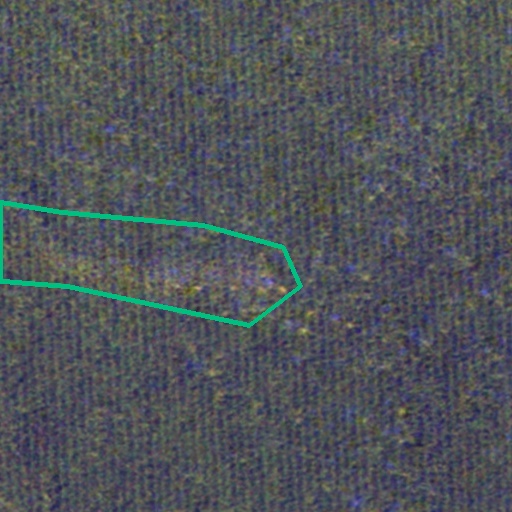}}%
        \imgbox{0.48}{\img{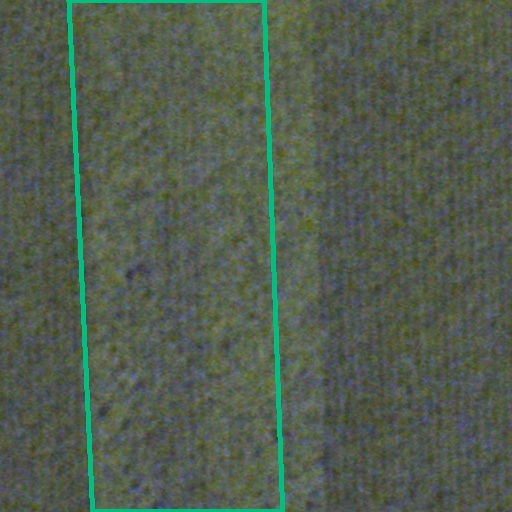}}
        \vspace{-0.5cm}
        \caption{Nutrient deficiency}
        \vspace{1cm}
    }%
    \imgbox{0.32}{
        \imgbox{0.48}{\img{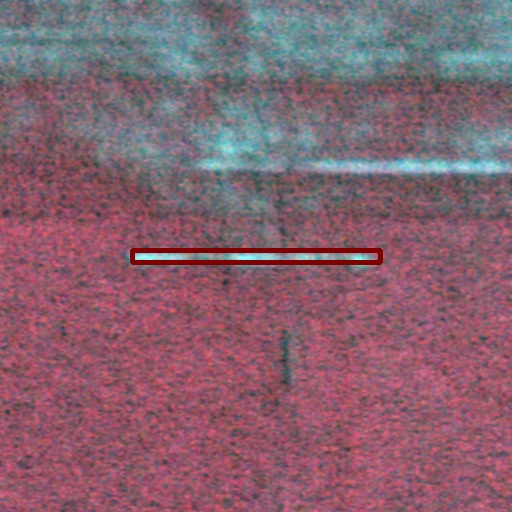}}%
        \imgbox{0.48}{\img{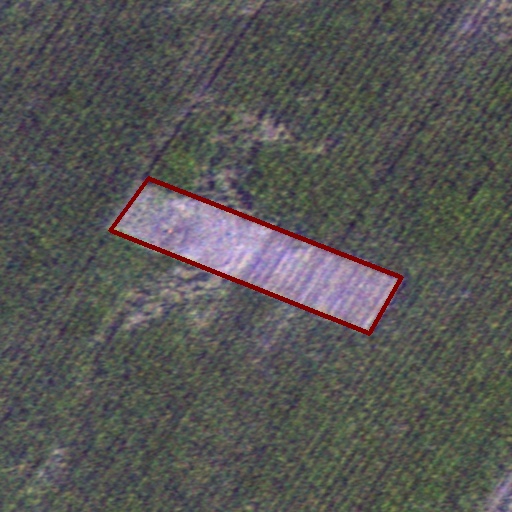}}
        \vspace{-0.5cm}
        \caption{Planter skip}
        \vspace{1cm}
    }%
    \imgbox{0.32}{
        \imgbox{0.48}{\img{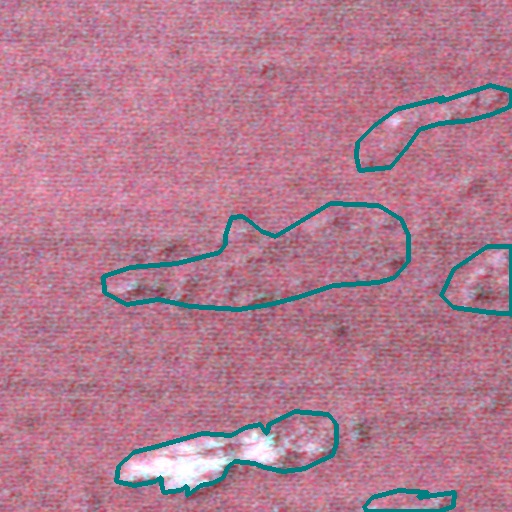}}%
        \imgbox{0.48}{\img{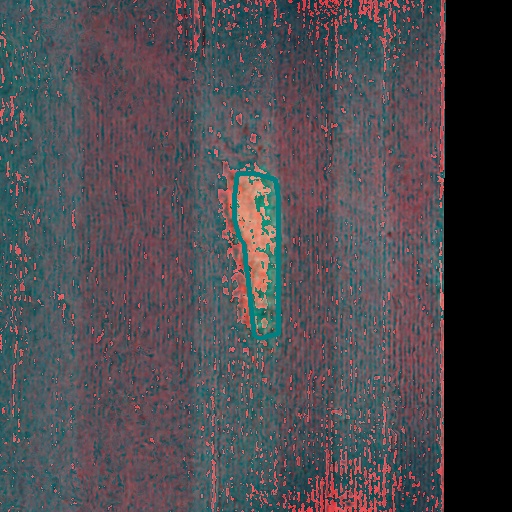}}
        \vspace{-0.5cm}
        \caption{Storm damage (not evaluated)}
        \vspace{1cm}
    }
    \imgbox{0.32}{
        \imgbox{0.48}{\img{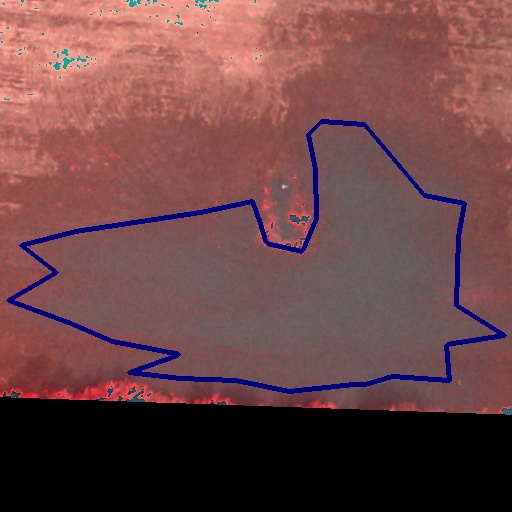}}%
        \imgbox{0.48}{\img{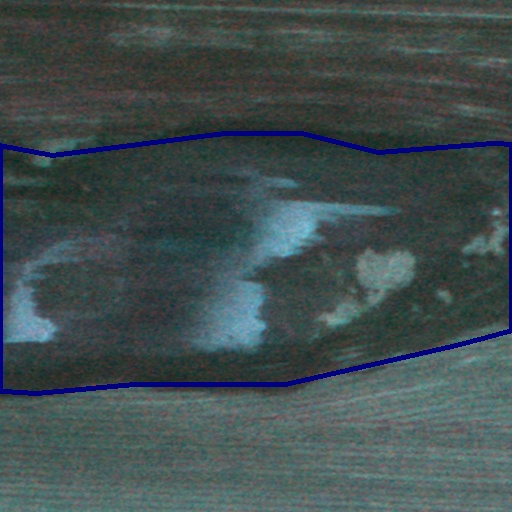}}
        \vspace{-0.5cm}
        \caption{Water}
        \vspace{0.42cm}
    }%
    \imgbox{0.32}{
        \imgbox{0.48}{\img{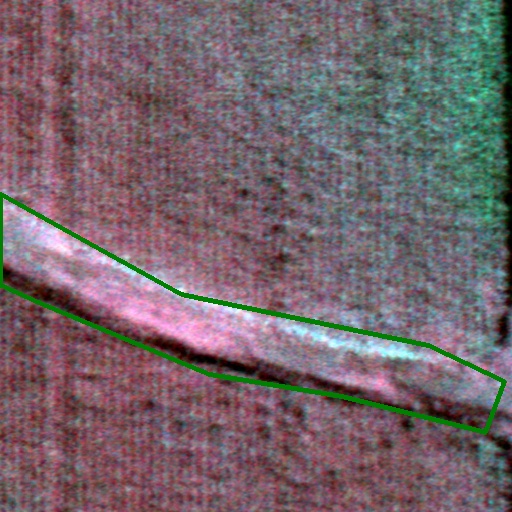}}%
        \imgbox{0.48}{\img{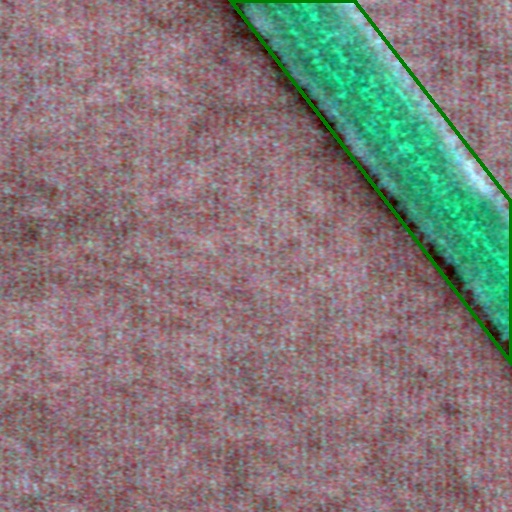}}
        \vspace{-0.5cm}
        \caption{Waterway}
        \vspace{0.42cm}
    }%
    \imgbox{0.32}{
        \imgbox{0.48}{\img{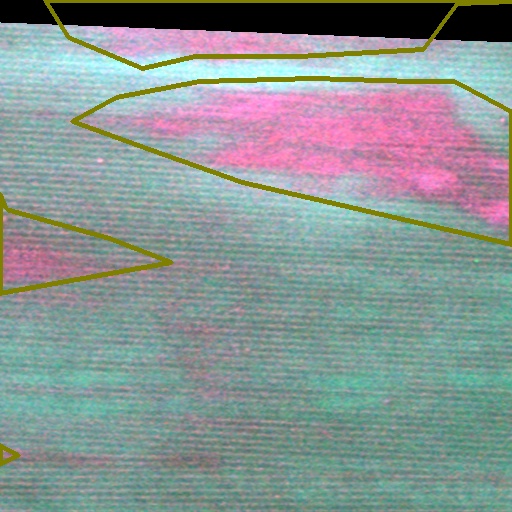}}%
        \imgbox{0.48}{\img{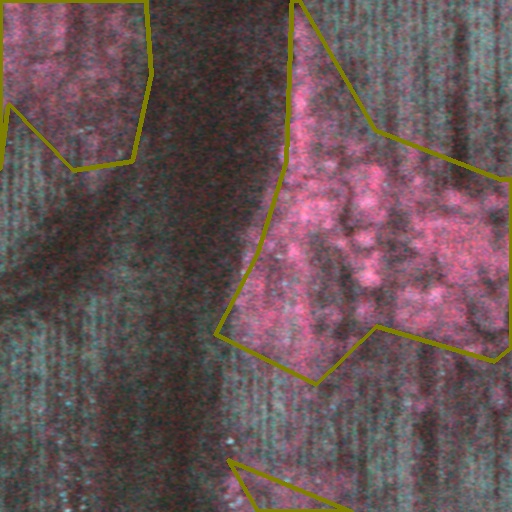}}
        \vspace{-0.5cm}
        \caption{Weed cluster}
        \vspace{0.42cm}
    }
\caption{For each annotation, top: RGB image; bottom: NRG image. Invalid regions have been blacked out. Note the extreme size and shape variations of some annotations. Note that images in our dataset can contain mutiple patterns, the visualizations above are chosen to best illustrate each pattern.
Images best viewed with color and zoomed in.}
\label{fig:crop_vis}
\end{figure*}

\begin{figure*}[ht!]
\centering
\begin{minipage}{0.31\linewidth}
    \centering
    \vspace{-3mm}
    \includegraphics[width=0.9\linewidth]{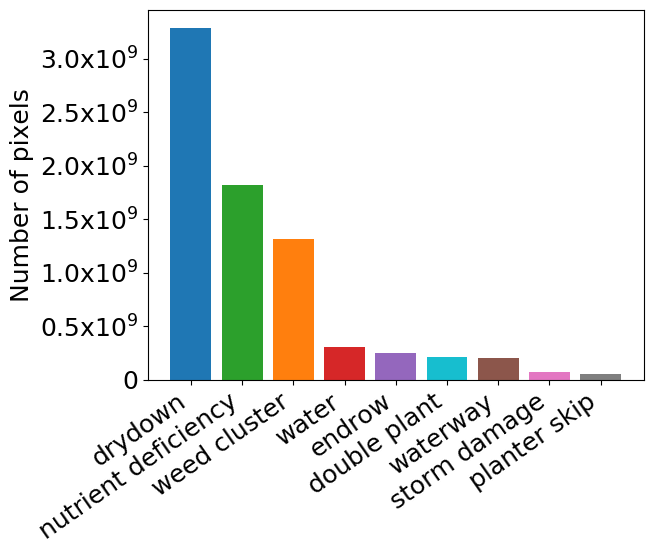}
    \vspace{-0.2cm}
    \caption{Total area of annotations for each class. Some categories occupy significantly larger areas than others, resulting in extreme class imbalance.}
    \label{fig:lbl_area}
\end{minipage}
\hspace{0.3cm}
\begin{minipage}{0.31\linewidth}
    \centering
    \includegraphics[width=0.9\linewidth]{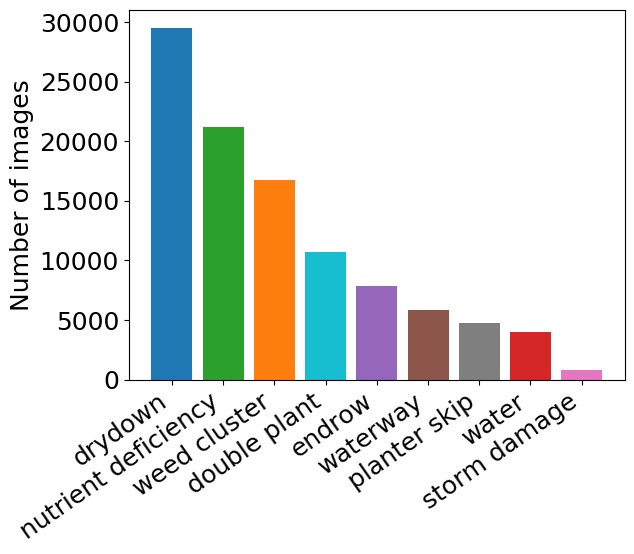}
    \vspace{-0.2cm}
    \caption{Number of images containing each anntoation class. A sudden drop in the number of storm damage samples indicate the difficulty for a model to recognize this pattern.}
    \label{fig:lbl_count}
\end{minipage}
\hspace{0.3cm}
\begin{minipage}{0.31\linewidth}
    \centering
    \vspace{-0.8cm}
    \includegraphics[width=0.9\linewidth]{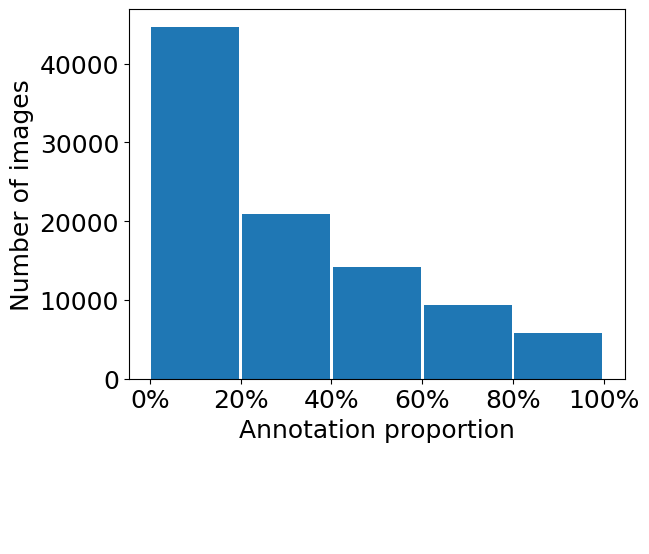}
    \vspace{-0.2cm}
    \caption{Percentages of annotated pixels in images. Some patterns are almost take up the entire image.}
    \label{fig:image_area}
\end{minipage}
\end{figure*}

\subsection{Dataset splitting}

We first randomly split the 3,432 farmland images with a 6/2/2 train/val/test ratio. We then assign each sampled image to the split of the farmland image they are cropped from. This guarantees that no cropped images from the same farmland will appear in multiple splits in the final dataset. The generated \dataset{} dataset thus contains 56,944/18,334/19,708 train/val/test images.

\section{Dataset Statistics}

\subsection{Annotation areas}
Field patterns have different shapes and sizes. For example, weed clusters can appear in either small patches or enormous regions, while double plant usually occur in small areas on the field. At the same time, these patterns also appear at different frequencies. Therefore, patterns that are large and more common occupy significantly larger areas than patterns that are smaller and relatively rare.

Figure~\ref{fig:lbl_area} shows the total number of pixels for each type of annotations in \dataset{}. We observe significantly more drydown, nutrient deficiency and weed cluster pixels than other categories in our dataset, which indicate extreme label imbalance across categories.

\subsection{Annotation counts}
The frequency at which a model observes a pattern during training determines the model's ability to recognize the same pattern during inference. It is therefore very important to understand the sample distribution for each of these field patterns in the  \dataset{} dataset.

Figure~\ref{fig:lbl_count} shows the number of images that contain each annotation category. While most annotations fall under a natural and smooth occurrence distribution, we observe a sudden drop of images containing storm damage patterns. The extreme scarcity of storm damage annotations would be problematic for model training. As a result, we ignore any storm damage annotations when performing evaluations.

\subsection{Annotation proportions}
As previously described, field patterns can vary dramatically in size. Correspondingly in \dataset{}, each generated image sample may also contain various proportions of annotations. We show in Figure~\ref{fig:image_area} that many images contain more than 50\% annotated pixels, some even occupy more than 80\% of the image. Training a model to segment large patterns can be difficult, since recognition of field patterns relies heavily on the contextual information of the surrounding field.

\section{Pilot Study on \dataset{}}
\label{sec:experiments}

\begin{table*}[t!]
\centering
\resizebox{0.95\textwidth}{!}{
    \renewcommand{\arraystretch}{1.1}
    \begin{tabular}{r|c|cccccccccc}
    \hline
    Model & mIoU (\%) & Background & \begin{tabular}[x]{@{}c@{}}Double\\plant\end{tabular} & Drydown & Endrow & \begin{tabular}[x]{@{}c@{}}Nutrient\\deficiency\end{tabular} & \begin{tabular}[x]{@{}c@{}}Planter\\skip\end{tabular} & Water & Waterway & \begin{tabular}[x]{@{}c@{}}Weed\\cluster\end{tabular} \\ \hline\hline
    DeepLabv3 (os=8) \cite{chen2017rethinking}  & 35.29          & 73.01          & 21.32          & 56.19          & 12.00          & 35.22          & 20.10          & 42.19          & 35.04          & 22.51          \\ 
    DeepLabv3+ (os=8) \cite{chen2018encoder}    & 37.95          & 72.76          & 21.94          & 56.80          & 16.88          & 34.18          & 18.80          & 61.98          & 35.25          & 22.98          \\ 
    DeepLabv3 (os=16) \cite{chen2017rethinking} & 41.66          & \textbf{74.45} & 25.77          & 57.91          & 19.15          & \textbf{39.40} & 24.25          & 72.35          & 36.42          & 25.24          \\ 
    DeepLabv3+ (os=16) \cite{chen2018encoder}   & 42.27          & 74.32          & 25.62          & \textbf{57.96} & 21.65          & 38.42          & 29.22          & 73.19          & \textbf{36.92} & 23.16          \\ \hline
    Ours                                        & \textbf{43.40} & 74.31          & \textbf{28.45} & 57.43          & \textbf{21.74} & 38.86          & \textbf{33.55} & \textbf{73.59} & 34.37          & \textbf{28.33} \\ \hline
    \end{tabular}
}
\vspace{-1mm}
\caption{mIoUs and class IoUs of modified semantic segmentation models and our proposed FPN-based model on \dataset{} \textbf{validation} set. Our model is customized for aerial agricultural images and perform better than all others.}
\label{table:val_result}
\vspace{-1mm}
\end{table*}
\begin{table*}[t!]
\centering
\resizebox{0.95\textwidth}{!}{
    \renewcommand{\arraystretch}{1.1}
    \begin{tabular}{r|c|ccccccccc}
    \hline
    Model & mIoU (\%) & Background & \begin{tabular}[x]{@{}c@{}}Double\\plant\end{tabular} & Drydown & Endrow & \begin{tabular}[x]{@{}c@{}}Nutrient\\deficiency\end{tabular} & \begin{tabular}[x]{@{}c@{}}Planter\\skip\end{tabular} & Water & Waterway & \begin{tabular}[x]{@{}c@{}}Weed\\cluster\end{tabular} \\ \hline\hline
    DeepLabv3 (os=8) \cite{chen2017rethinking}  & 32.18          & 70.42          & 21.51          & 50.97          & 12.60          & 39.37          & 20.37          & 15.69          & 33.71          & 24.98          \\ 
    DeepLabv3+ (os=8) \cite{chen2018encoder}    & 39.05          & 70.99          & 19.67          & 50.89          & 19.50          & 41.32          & 24.42          & 62.25          & 34.14          & 28.27          \\ 
    DeepLabv3 (os=16) \cite{chen2017rethinking} & 42.22          & \textbf{72.73} & 25.15          & \textbf{53.62} & 20.99          & 43.95          & 24.57          & 70.42          & 38.63          & 29.91          \\ 
    DeepLabv3+ (os=16) \cite{chen2018encoder}   & 42.42          & 72.50          & 25.99          & 53.57          & 24.10          & \textbf{44.15} & 24.39          & 70.33          & 37.91          & 28.81          \\ \hline
    Ours                                        & \textbf{43.66} & 72.55          & \textbf{27.88} & 52.32          & \textbf{24.43} & 43.79          & \textbf{30.95} & \textbf{71.33} & \textbf{38.81} & \textbf{30.87} \\ \hline
    \end{tabular}
}
\vspace{-1mm}
\caption{mIoUs and class IoUs of semantic segmentation models and our proposed model on \dataset{} \textbf{test} set. The results are consistent with the validation set, where our model outperforms common object semantic segmentation models.}
\label{table:test_result}
\vspace{-2mm}
\end{table*}


\subsection{Baseline models}

There are many popular models for semantic segmentation on common object datasets. For example, U-Net~\cite{ronneberger2015u} is a light-weight model that leverages an encoder-decoder architecture for pixel-wise classification. PSPNet~\cite{zhao2017pyramid} uses spatial pooling at multiple resolutions to gather global information. DeepLab~\cite{chen2014semantic, chen2017deeplab, chen2017rethinking, chen2018encoder} is a well-known series of deep learning models that use atrous convolutions for semantic segmentation. More recently, many new methods have been proposed and achieve state-of-the-art results on CityScapes benchmark. For example, SPGNet~\cite{cheng2019spgnet} proposes a Semantic Prediction Guidance (SPG) module which learns to re-weight the local features through the guidance from pixel-wise semantic prediction, and \cite{huang2019ccnet} proposes Criss-Cross Network (CCNet) for obtaining better contextual information in a more effective and efficient way. In our experiments, we perform comparative evaluations on the \dataset{} dataset using DeepLabV3 and DeepLabV3+, which are two well-performing models across several semantic segmentation datasets. We also propose a specialized FPN-based model that outperforms these two milestones in \dataset{}.

To couple with \dataset{}, we make minor modifications on the existing DeepLabV3 and DeepLabV3+ architectures. Since \dataset{} contains NRGB images, we duplicate the weights corresponding to the Red channel of the pretrained convolution layer. This gives a convolution layer with four input channels in the backbone.

\subsection{The proposed FPN-based model}

In our FPN-based model, the encoder of the FPN is a ResNet~\cite{he2016deep}. We retain the first three residual blocks of the ResNet, and we change the last residual block (layer4) into a dilated residual block with rate=4. The modified block shares the same structure with the Deeplab series~\cite{chen2014semantic, chen2017deeplab, chen2017rethinking, chen2018encoder}. We implement the lateral connections in the FPN decoder using two $3\times 3$ and one $1 \times 1$ convolution layers. Each of the two $3\times 3$ convolution layers is followed by a batch normalization layer (BN) and a leaky ReLU activation with a negative slope of 0.01. The last $1 \times 1$ convolution layer does not contain bias units. For the upsampling modules, instead of bilinear interpolation, we use a deconvolution layer with kernel size=3, stride=2 and padding=1, followed by a BN layer, leaky ReLU activation and another $1 \times 1$ convolution layer without bias. The output from each lateral connection and the corresponding upsampling module are added together, the output is then passed through two more $3 \times 3$ convolution layers with BN and leaky ReLU. Lastly, outputs from all pyramid levels are upsampled to the highest pyramid resolution using bilinear interpolation and are then concatenated. The result is passed to a $1 \times 1$ convolution layer with bias units to predict the final semantic map.

\subsection{Training details}

We use backbone models pretrained on ImageNet in all our experiments. We train each model for 25,000 iterations with a batch size of 40 on four RTX 2080Ti GPUs. We use SGD with a base learning rate of 0.01 and a weight decay of $5\times10^{-4}$. Within the 25,000 iterations, we first warm-up the training for 1,000 iterations~\cite{linear_warmup}, where the learning rate linearly grows from 0 to 0.01. We then train for 7,000 iterations with a constant learning rate of 0.01. We finally decrease the learning rate back to 0 with the ``poly'' rule \cite{chen2017deeplab} in the remaining 17,000 iterations. 

Table~\ref{table:val_result} and Table~\ref{table:test_result} show the validation and test set results of DeepLabV3 and DeepLabV3+ with different output strides and our proposed FPN-based model. Our model consistently outperforms these semantic segmentation models in \dataset{}. Hence, in the following experiments, we will use our FPN-based model for comparison studies.

\subsection{Multi-spectral data and model complexity}

One major study of our work is the effectiveness of training multi-spectral data in image recognition. \dataset{} consists of NIR-Red-Green-Blue (NRGB) images, which is beyond many conventional image recognition tasks. Therefore, we investigate the differences in performance between semantic segmentation from multi-spectral images, including NRG and NRGB images, and regular RGB images.

We simultaneously investigate the impact of using models with different complexities. Specifically, we train our FPN-based model with ResNet-50 and ResNet-101 as backbone. We evaluate combinations of multi-spectral images and various backbones and report the results in Table~\ref{table:multi_spectral_backbone}.

\begin{table}[t!]
\centering
\def\arraystretch{1.3}
\resizebox{0.9\linewidth}{!}{
    \begin{tabular}{c|c|c|c}
    \hline
    Backbone   & Channels & Val mIoU (\%)  & Test mIoU (\%) \\ \hline\hline
    ResNet-50  & RGB     & 39.28          & 38.26          \\
    ResNet-50  & NRG     & 42.89          & 41.34          \\
    ResNet-50  & NRGB    & 42.16          & 41.82          \\ \hline
    ResNet-101 & RGB     & 40.48          & 39.63          \\
    ResNet-101 & NRG     & 42.25          & 40.05          \\
    ResNet-101 & NRGB    & \textbf{43.40} & \textbf{43.66} \\ \hline    
    \end{tabular}
}
\caption{mIoUs using our proposed model with various ResNet backbones and image channels.}
\label{table:multi_spectral_backbone}
\end{table}

\subsection{Multi-scale data}

Aerial farmland images contain annotations with vastly different sizes. As a result, models trained from images at different scales can result in significantly different performances. In order to justify our choice of using $512\times512$ windows to construct the \dataset{} dataset, we additionally generate two versions of the dataset with different window sizes. The first version (\dataset{}-1024) uses $1024\times1024$ windows to crop annotations. The second version (\dataset{}-MS) uses three window sizes: $1024\times1024$, $1536\times1536$ and $2048\times2048$. 

In \dataset{}-MS, images are cropped with the smallest window size that completely encloses the annotation. If an annotation exceeds $2048\times2048$ pixels, we again use the sliding window cropping method to generate multiple sub-samples. We use \dataset{}-MS to evaluate if retaining the integrity of large annotations helps to improve performances. Note that this is different from conventional multi-scale inference used in common object image segmentation, since in \dataset{}-MS the images are of different sizes.

We cross-evaluate models trained on each dataset version with all three versions. Results in Table~\ref{table:multi_scale} show that the model trained on the proposed \dataset{} dataset with a $512\times512$ window size is the most stable and performs the best, thus justifying our dataset with the chosen image sampling method.

\section{Discussion}

We would like to highlight the use of \dataset{} to tackle the following crucial tasks:

\begin{itemize}
    \item \textbf{Agriculture images beyond RGB:} Deep convolutional neural networks (DCNN) are channel-wise expandable by nature. Yet few datasets promote in-depth research on such capability. We have demonstrated that aerial agricultural semantic segmentation is more effective using NRGB images rather than just RGB images. Future versions of \dataset{} will also include thermal images, soil maps and topographic maps. Therefore, further studies in multi-spectral agriculture images are within our expectation.

    \item \textbf{Transfer learning:} Our segmentation task induces an uncommon type of transfer learning, where a model pretrained on RGB images of common objects is transferred to multi-spectral agricultural images. Although the gap between the source and target domain is tremendous, our experiments show that transfer learning remains an effective way of learning to recognize field patterns. Similar types of transfer learning are not regularly seen, but they are expected to become more popularized with \dataset{}. The effectiveness of fine-tuning can be further explored, such as channel expansion in convolution layers and domain adaptation from common objects to agricultural patterns.
    
    \item \textbf{Learning from extreme image sizes:} The current version of \dataset{} provides a pilot study of aerial agricultural pattern recognition with conventional image sizes. However, our multi-scale experiments show that there is still much to explore in effectively leveraging large-scale aerial images for improved performance. Using \dataset{} as a starting point, we hope to initiate related research on visual recognition tasks that are generalizable to extremely large aerial farmland images. We envision future work in this direction to enable large-scale image analysis as a whole.
    
\end{itemize}

    
    

    
    

\begin{table}[t!]
\centering
\def\arraystretch{1.2}
\resizebox{0.95\linewidth}{!}{
    \begin{tabular}{c|c|c|c|c|c|c|c}
    \hline
    \multicolumn{2}{c|}{} & \multicolumn{3}{c|}{Val mIoU (\%)} & \multicolumn{3}{c}{Test mIoU (\%)} \\ \cline{3-8}
    \multicolumn{2}{c|}{} & 512 & 1024 & MS & 512 & 1024 & MS \\ \hline\hline
    \multirow{3}{*}{\rotatebox[origin=c]{90}{Train}} 
    & 512  & \textbf{43.40} & \textbf{39.44} & \textbf{37.64} & \textbf{43.66} & \textbf{39.68} & \textbf{37.27} \\ 
    & 1024 & 36.33 & 34.37 & 36.16 & 35.01 & 35.27 & 35.87 \\ 
    & MS   & 34.16 & 32.45 & 35.67 & 31.17 & 30.72 & 35.77 \\ \hline
    \end{tabular}
}
\caption{mIoUs of our model trained and tested on different \dataset{} versions. 512: the proposed \dataset{} dataset, 1024: \dataset{}-1024, MS: \dataset{}-MS. The model trained on the proposed dataset yields the best results across all versions.}
\label{table:multi_scale}
\end{table}

\section{Conclusion}
We introduce \dataset{}, an aerial agricultural semantic segmentation dataset. We capture extremely large farmland images and provide multiple field pattern annotations. This dataset poses new challenges in agricultural semantic segmentation from aerial images. As a baseline, we provide a pilot study on \dataset{} using well-known off-the-shelf semantic segmentation models and our specialized one.

In later versions, \dataset{} will include more field images and patterns, as well as more image modalities, such as thermal images, soil maps and topographic maps. This would make \dataset{} an even more standardized and inclusive aerial agricultural dataset. We hope this dataset will encourage more work on improving visual recognition methods for agriculture, particularly on large-scale, multi-channel aerial farmland semantic segmentation.

{\small
\bibliographystyle{ieee_fullname}
\bibliography{egbib}
}

\end{document}